\documentclass[sn-mathphys,iicol]{sn-jnl}

\usepackage{amsmath,amsfonts,bm}

\def\eqref#1{equation~\ref{#1}}

\def\1{\bm{1}}

\DeclareMathAlphabet{\mathsfit}{\encodingdefault}{\sfdefault}{m}{sl}
\SetMathAlphabet{\mathsfit}{bold}{\encodingdefault}{\sfdefault}{bx}{n}

\newcommand{\E}{\mathbb{E}}

\newcommand{\bbR}{\mathbb{R}}

\usepackage{times,url,booktabs,microtype,graphicx,bbm,verbatim,amsthm,amssymb,amsmath,enumitem,pifont,makecell,multirow,mdframed,xcolor,colortbl,rotating,wrapfig,resizegather,algorithm}

\usepackage{algpseudocode}
\usepackage{caption}
\usepackage{subcaption}

\newcommand{\bmx}{{\bm x}}

\newcommand{\bmz}{{\bm z}}

\usepackage{commath}

\jyear{2021}%

\theoremstyle{thmstyleone}%
\theoremstyle{thmstyletwo}%

\theoremstyle{thmstylethree}%

\begin{document}

\title[Sparse-to-Sparse GAN Training Without Sacrificing Performance]{Don't Be So Dense: Sparse-to-Sparse GAN Training Without Sacrificing Performance}


\author[1]{\fnm{Shiwei} \sur{Liu}}\email{s.liu3@tue.nl}

\author[2]{\fnm{Yuesong} \sur{Tian}}\email{
tianys163@gmail.com}

\author[3]{\fnm{Tianlong} \sur{Chen}}\email{
tianlong.chen@utexas.edu}

\author*[4]{\fnm{Li} \sur{Shen}}\email{
mathshenli@gmail.com}

\affil[1]{\orgdiv{Computer Science}, \orgname{Eindhoven University of Technology}, \orgaddress{\street{Groene Loper 5}, \city{Eindhoven}, \postcode{5612 AZ}, \country{the Netherlands}}}

\affil[2]{\orgdiv{College of Biomedical Engineering and Instrument
Science}, \orgname{Zhejiang University}, \orgaddress{\street{Zheda Road}, \city{Hangzhou}, \postcode{310000}, \state{Zhejiang}, \country{China}}}

\affil[3]{\orgdiv{Electrical and Computer Engineering}, \orgname{University of Texas at Austin}, \orgaddress{\street{2501 Speedway}, \city{Austin}, \postcode{78712}, \state{Texas}, \country{United States}}}

\affil*[4]{\orgname{JD Explore Academy}, \orgaddress{\city{Beijing}, \country{China}}}


\abstract{Generative adversarial networks (GANs) have received an upsurging interest since being proposed due to the high quality of the generated data. While achieving increasingly impressive results, the resource demands associated with the large model size hinders the usage of GANs in resource-limited scenarios. For inference, the existing model compression techniques can reduce the model complexity with comparable performance. However, the training efficiency of GANs has less been explored due to the fragile training process of GANs. In this paper, we, for the first time, explore the possibility of directly training sparse GAN from scratch without involving any dense or pre-training steps. Even more unconventionally, our proposed method enables directly training \textit{sparse unbalanced GANs} with an extremely sparse generator from scratch. Instead of training full GANs, we start with sparse GANs and dynamically explore the parameter space spanned over the generator throughout training. Such a sparse-to-sparse training procedure enhances the capacity of the highly sparse generator progressively while sticking to a fixed small parameter budget with appealing training and inference efficiency gains. Extensive experiments with modern GAN architectures validate the effectiveness of our method. Our sparsified GANs, trained from scratch in one single run, are able to outperform the ones learned by expensive iterative pruning and re-training. Perhaps most importantly, we find instead of inheriting parameters from expensive pre-trained GANs, directly training sparse GANs from scratch can be a much more efficient solution. For example, only training with a 80\% sparse generator and a 70\% sparse discriminator, our method can achieve even better performance than the dense BigGAN.}

\keywords{sparse GAN training, sparse unbalance GAN, dynamic sparsity, sparse training, generative adversarial networks}



\maketitle

\section{Introduction}\label{sec1}

The past decade has witnessed impressive results achieved by generative adversarial networks (GANs)~\cite{NIPS2014_5ca3e9b1,zhu2017unpaired,arjovsky2017wasserstein,miyato2018spectral,miyato2018cgans,brock2018large,karras2017progressive,karras2019style,karras2020analyzing}. In concert with the improved quality of the generated data, the training and inference costs of the state-of-the-art GANs have also exploded, curbing the application of GANs in edge devices. Reducing computational costs and memory requirements is of importance for many GAN-based applications. 

Prior works utilize model compression techniques, such as pruning~\cite{shu2019co}, distillation~\cite{li2020gan,chen2020distilling,wang2020gan}, quantization~\cite{wang2019qgan}, and lottery tickets hypothesis (LTH)~\cite{frankle2018lottery,chen2021gans,chen2021ultra} to produce an efficient generator with competitive performance. While increasingly efficient, the existing techniques are not designed to accelerate training as they either operate on fully pre-trained GANs or require the over-parameterized dense GANS to be stored or updated during training. As the resource demands associated with training increase quickly~\cite{strubell2019energy}, such highly over-parameterized dependency may lead to financial and environmental concerns~\cite{patterson2021carbon}. For example, while the sparse GANs (winning tickets) learned by LTH can match the performance of the dense model, the identification of these winning tickets involves accomplishing the costly train-prune-retrain process many times, resulting in much greater overall FLOPs than training a dense GAN model.

Instead of inheriting knowledge from pre-trained GANs, it is more desirable to train \textit{intrinsically sparse GANs from scratch} in an end-to-end way (sparse-to-sparse training). What's more, training sparse unbalanced\footnote{Sparse unbalanced GANs refer to the scenarios where the sparsities of generators and discriminators are not well-matched.} GANs consisting of an extremely sparse generator and a much denser discriminator are arguably more appealing, as such an unbalanced training process will yield extremely sparse generators, very desirable for inference. However, this tantalizing perspective has never been fulfilled due to several daunting challenges: (1) just like most other deep neural networks~\cite{mocanu2016topological,evci2019difficulty}, naively training sparse GANs from scratch without pre-training knowledge typically leads to unsatisfactory performance~\cite{yu2020self}. Therefore, it remains mysterious whether we can train extremely sparse GANs (even if balanced) to match the performance of the dense equivalents; (2) it is well-known that dense GANs suffer from notorious training instability~\cite{berthelot2017began,ham2020unbalanced}. The large sparsity unbalance between generators and discriminators will only make it worse. This naturally raises a question: \textit{can we train a sparse unbalanced GAN model with an extremely sparse generator from scratch without sacrificing performance?}

\begin{figure*}[t!]

\begin{center}
\centerline{\includegraphics[width=0.7\textwidth]{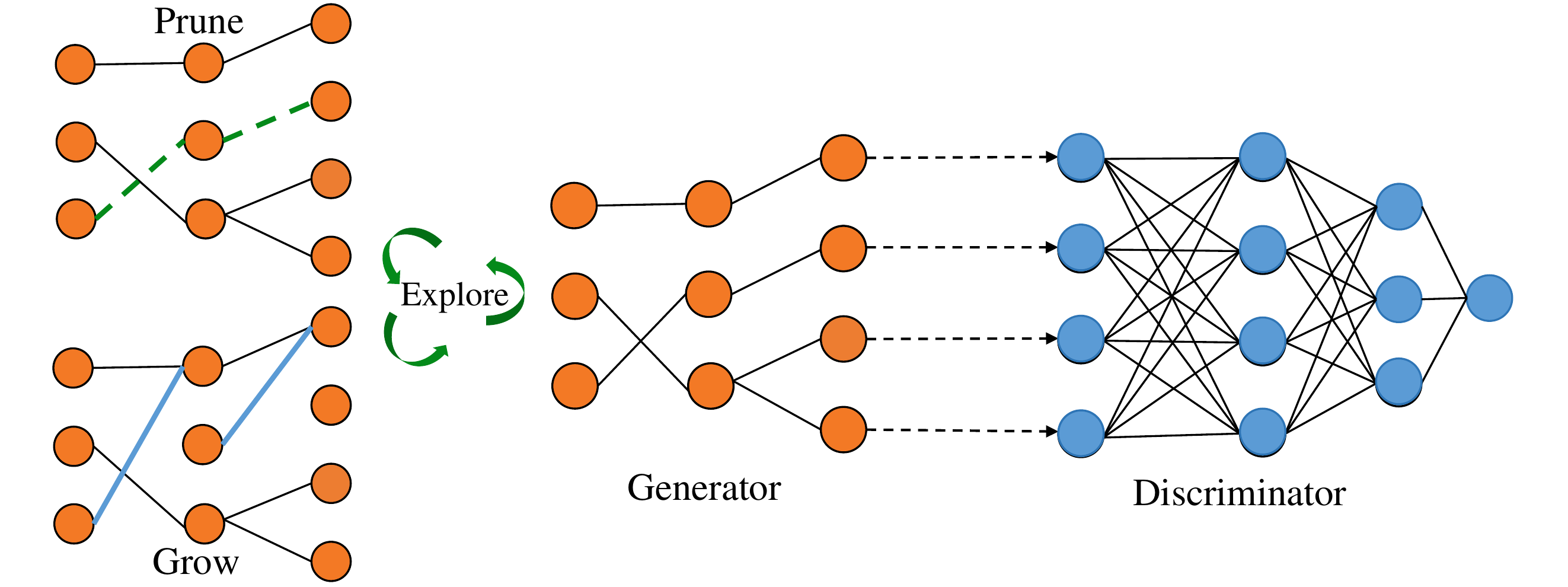}}
\caption{STU-GAN directly trains sparsity-unbalanced GANs from scratch with an extremely sparse generator and a much denser discriminator. The training instability is mitigated by periodically exploring parameters for better sparse connectivities with a prune-and-redistribute scheme during training. }
\label{Fig:STU_GAN}
\end{center}
\end{figure*}

 
Intuitively, the extremely sparse generator is too weak to generate high-quality data that can fool the discriminator. To maintain the benign competition of the minimax game behind GANs, we need to improve the expressibility of the sparse generator. One intuitive solution is to inherit weights or ``lucky'' initializations from pre-trained GANs~\cite{chen2021gans}, which unfortunately involves the expensive pre-training.

In this paper, we attempt to address this  research question by proposing a novel approach termed Sparse Unbalanced GAN (STU-GAN). Instead of training highly over-parameterized, dense GANs, STU-GAN is able to train intrinsically sparse GANs comprising an extremely sparse generator and a much denser discriminator from scratch, without involving any pre-training or dense training steps. The key of STU-GAN lies in the parameter exploration of the sparse generator, which substantially improves the expressibility of the sparse generator during the course of training, while maintaining a fixed number of parameters. STU-GAN periodically explores the parameter space and search for better sparse connections from which the sparse generator constructed has an increasingly stronger generative ability. Thanks to the cost-free parameter exploration, STU-GAN significantly mitigates the training instability of sparse unbalanced GANs, ending up with an extreme sparse generator for inference. We summarize our main contributions below:



\begin{itemize}
\item We first study the sensitivity of the two most common sparsity-inducing techniques, i.e., fine-tuning after pruning and sparse training from scratch, to the scenario with sparsity unbalance. We empirically find they all severely suffer from the unbalanced sparsity allocation with extremely sparse generators, indicating challenges of sparse unbalanced GAN Training.

\item To improve the trainability of sparse unbalanced GANs, we propose an approach termed Sparse Training Unbalanced GAN (STU-GAN). STU-GAN directly trains a sparse unbalanced GAN with an \underline{extremely sparse generator} and a much denser discriminator from scratch without involving any expensive dense or pre-training steps. By thoroughly exploring the parameter space spanned over the sparse generator, STU-GAN improves its expressibility, and hence stabilizes the atypical training procedure of sparse unbalanced GANs while sticking to a fixed small parameter budget.

\item Extensive experiments are conducted with BigGAN~\cite{brock2018large} on CIFAR-10 and ImageNet, SNGAN~\cite{miyato2018spectral} on CIFAR-10. The consistent performance improvement over the existing techniques verifies the effectiveness of our proposal. Specifically, STU-GAN outperforms dense BigGAN on CIFAR-10 only with a 80\% sparse generator and a 70\% sparse discriminator, while being end-to-end trainable.

\end{itemize}

\section{Related Work}
\label{sec:related_work}
\subsection{GAN Compression} While enjoying success in image generation and translation tasks~\cite{karras2017progressive,chen2018cartoongan,jing2019neural,gui2020review}, generative adversarial networks (GANs), like other deep neural networks, also inherit the undesirable property -- high computational complexity and memory requirements. To compress GANs, Han et al.~\cite{shu2019co}  proposed a co-evolutionary pruning algorithm to simultaneously pruning redundant filters in both generators. QGAN was proposed by Wang et al.~\cite{wang2019qgan}, which can quantize GANs to 2-bit or 1-bit still preserving comparable quality. Distillation was also used by Li et al.~\cite{li2020gan} to enhance the compressed discriminator with a pre-trained GAN model. Wang et al.~\cite{wang2020gan} moved one step further and combined the above-mentioned three techniques into a unified framework. A pre-trained discriminator is used by~\cite{yu2020self} to supervise the training of a compressed generator, achieving compelling results. Very recently, the authors of ~\cite{chen2021gans} extended LTH into GANs, verifying the existence of winning tickets in deep GANs. The existing works on GAN compression all require training  an over-parameterized GAN model in advance, not designed for training efficiency. In contrast, our method directly trains a sparse GAN in an end-to-end way, bringing efficiency gains to both training and inference. 

\subsection{Sparse-to-Sparse Training}
Recently, researchers began to investigate techniques to train intrinsically sparse neural networks from scratch (sparse-to-sparse training). According to whether the sparse connectivity dynamically changes during training,  sparse-to-sparse training can be divided into static sparse training (SST) and dynamic sparse training (DST).

\textbf{Static sparse training.} SST represents a class of methods that aim to train a sparse subnetwork with a fixed sparse connectivity pattern during the course of training. The most naive approach is pruning each layer uniformly with the same pruning ratio, i.e., uniform pruning~\cite{mariet2015diversity,he2017channel,suau2019network,gale2019state}. Mocanu et al.~\cite{Mocanu2016xbm} proposed a non-uniform and scale-free topology, showing better performance than the dense counterpart when applied to restricted Boltzmann machines (RBMs). Later, expander graphs were introduced to build sparse CNNs and showed comparable performance against the corresponding dense CNNs~\cite{prabhu2018deep,kepner2019radix}. While not initially designed for SST, \textit{Erd{\H{o}}s-R{\'e}nyi} (ER)~\cite{mocanu2018scalable} and \textit{Erd{\H{o}}s-R{\'e}nyi-Kernel} (ERK)~\cite{evci2020rigging} are two advanced layer-wise sparsities introduced from the field of graph theory with strong results. 

\textbf{Dynamic sparse training.} Mocanu et al.~\cite{mocanu2018scalable} first proposed sparse evolutionary training (SET) that uses a simple prune-and-regrow scheme to update the sparse connectivity, demonstrating better performance than training with static connectivity~\cite{mocanu2016topological,gale2019state}. Following this, weights redistribution are introduced to search for better layer-wise sparsity ratios~\cite{mostafa2019parameter,dettmers2019sparse,liu2021selfish}. While most DST methods use magnitude pruning to remove parameters, there is a large discrepancy between their redistribution criteria. Gradient-based regrowth e.g., momentum~\cite{dettmers2019sparse} and gradient~\cite{evci2020rigging} shows strong results in convolutional neural networks, whereas random regrowth outperforms the former in language modeling~\cite{dietrich2021towards}. Very recently, Liu et al.~\cite{liu2021we} conjectured that the expressibility of sparse training is highly correlated with the overall number of explored parameters during training. Inspired by the observations in~\cite{liu2021we}, we leverage parameter exploration in the extreme sparse generators to address the severe disequilibrium problem in sparse unbalanced GANs, for the appealing efficiency from training to inference. 

\section{The Difficulty of Sparse Unbalanced GAN Training}

\subsection{Preliminary and Setups} Generative Adversarial Networks (GANs) consist of a Generator $G(z, \bm \theta_G)$ and a Discriminator $D(x,  \bm \theta_D)$. The goal of $G(z, \bm \theta_G)$ is to map a sample $z$ from a random distribution $p(\bmz)$ to the data distribution $q_{ data}(\bmx)$ whereas the goal of $D(x, \bm \theta_D)$ is to determine whether a sample $x$ belongs to the data distribution. Formally, the original dense GANs objective from~\cite{NIPS2014_5ca3e9b1} is given as follows: 
\begin{equation}
\begin{split}
	 \min_{\bm \theta_G} \max_{\bm \theta_D}( \E_{x\sim q_{\rm data}({\bm x})} [ \log D({ x, \bm \theta_D})] \\
	 + \E_{{\bm z}\sim p({\bm z})} [\log(1-D(G({ z},\bm \theta_G)))]) \label{eq:advloss}
\end{split}
\end{equation}
where $\bm z \in \bbR^{d_z}$ is a latent variable drawn from a random distribution $p(\bm z)$. Consequently, the objective of sparse GANs can be formaltated as:
\begin{equation}
\begin{split}
	\min_{\bm \theta_{s_G}} \max_{\bm \theta_{s_D}}(\E_{x\sim q_{\rm data}({\bm x})} [ \log D({ x}, {\bm \theta_{s_{D}}})]  \\
	+ \E_{{\bm z}\sim p({\bm z})} [\log(1-\!D(G({ z}, {\bm \theta_{s_{G}}})))])
	\label{eq:sparseadvloss}
\end{split}
\end{equation}
where the sparse generator $G({  z}, {\bm \theta_{s_{G}}})$ and sparse discriminator $D({ x}, {\bm \theta_{s_{D}}})$ is parameterized by a fraction of parameters (subnetworks) $\bm \theta_{s_{G}}$ and $\bm \theta_{s_{D}}$, respectively. We define sparsity (i.e., fraction of zeros) of $\bm \theta_{s_{G}}$ and $\bm \theta_{s_{D}}$ as $s_{G}=1-\frac{\|\bm \theta_{s_{G}}\|_0}{\|\bm \theta_{G}\|_0}$ and $s_{D}=1-\frac{\|\bm \theta_{s_{D}}\|_0}{\|\bm \theta_{D}\|_0}$, individually, where $\|\cdot\|_0$ is the $\ell_0$-norm. We consider unstructured sparsity (individual weights are removed from a network) in this paper, not only due to its promising ability to preserve performance even at extreme sparsities~\cite{frankle2018lottery,evci2020rigging} but also the increasing support for sparse operations on the practical hardware~\cite{gale2020sparse,onemillionneurons,nvidia2020,zhou2021learning}. 

We first study the effect of sparsity unbalance on two sparsity-inducing techniques applied to GANs, (i) pruning and fine-tuning, and (ii) sparse training from scratch. More specifically, we report Fr\'echet Inception Distance (FID) achieved by these two methods under two scenarios: (1) only $G(z, \bm \theta_G)$ is sparsified; (2) both $G(z, \bm \theta_G)$ and $D(x, \bm \theta_D)$ are sparsified. The latter has barely been studied due to the fact that existing methods mainly focus on accelerating inference, no need to prune $D(x, \bm \theta_D)$.

\begin{figure*}[t]
\begin{center}
\centerline{\includegraphics[width=0.9\textwidth]{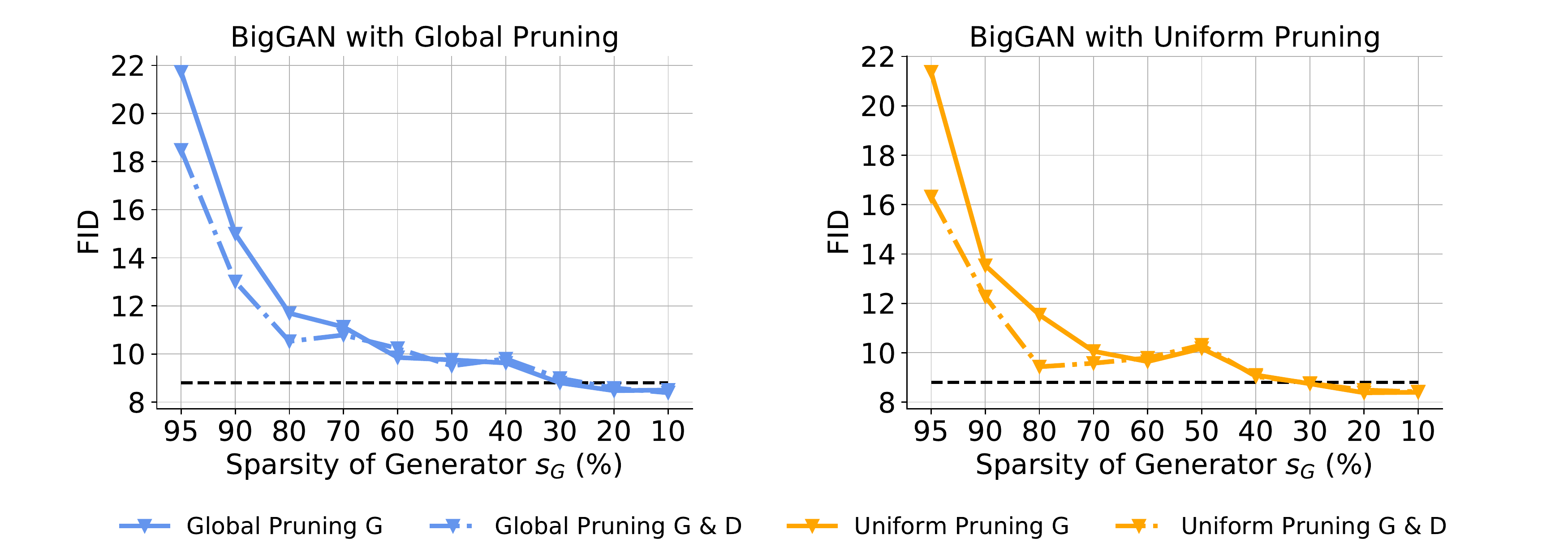}}
\caption{Effect of sparsity unbalance on pruning and fine-tuning. Experiments are conducted with BigGAN on CIFAR-10. Higher $s_G$ refers to fewer parameters remaining in generators. ``Global Pruning'' refers to pruning weights across layers and uniform pruning refers to pruning layer-wisely. ``Pruning G'' means  only pruning the generator while keeping the discriminator dense. ``Pruning G \& D'' means  pruning the generator and the discriminator together.}
\label{Fig:fine-tuning}
\end{center}
\end{figure*}

\begin{figure*}[ht]
\begin{center}
\centerline{\includegraphics[width=0.9\textwidth]{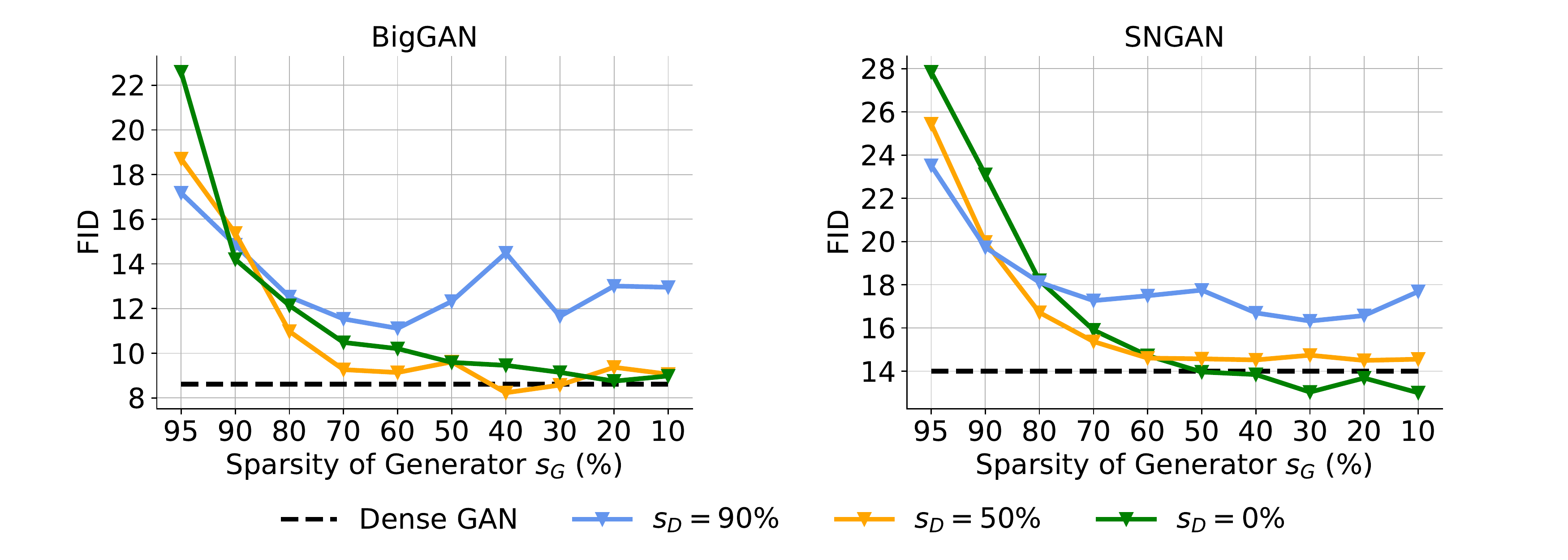}}
\caption{Effect of sparsity unbalance on GAN training. Higher $s_G$ and $s_D$ refer to fewer parameters remaining in the corresponding networks. Global pruning refers to pruning weights across layers and uniform pruning refers to pruning layer-wisely.}
\label{Fig:static_sparse}
\end{center}
\end{figure*}

\subsection{Pruning and Fine-tuning} Pruning and fine-tuning, as the most common pruning technique, prunes the pre-trained model to the target sparsity first and then trains it for further $t$ epochs. In this paper, we choose one-shot magnitude pruning~\cite{han2015learning,frankle2018lottery}, which removes the weights with the smallest magnitude in one iteration of pruning. Doing so makes a
fair comparison to sparse training, which does not involve any iterative re-training. Magnitude pruning is performed both uniformly (i.e., weights are removed layer by layer) and globally (i.e., weights are removed across layers). After pruning, the pruned models are further fine-tuned for the same number of epochs as the dense GANs. Intuitively, we should prune and fine-tune the generator together with the discriminator due to the instability of GAN training and hence fine-tuning. Figure~\ref{Fig:fine-tuning} shows that pruning and fine-tuning is indeed sensitive to sparsity unbalance. As expected, pruning and fine-tuning $G(z, \bm \theta_G)$ and $D(x, \bm \theta_D)$ together achieve lower FID than only pruning $G(z, \bm \theta_G)$, indicating the vital role of the balanced sparsity in pruning and fine-tuning.


\subsection{Sparse Training from Scratch} To largely trim down both the training and inference complexity, we prefer training an unbalanced sparse GAN in which $s_{G} > s_{D}$. However, balancing the training of dense GANs is already a challenge~\cite{berthelot2017began}. Training a sparse unbalanced GAN is even more daunting. To comprehensively understand the effect of sparsity unbalance on GAN training, we directly train sparse GANs without any parameter exploration (Static Sparse GAN) at various combinations of $s_D$ and $s_G$.  Concretely, we fix $s_D\in[0\%, 50\%, 90\%]$ and vary $s_G$ from 10\% to 95\%. As shown in Figure~\ref{Fig:static_sparse}, the balance of sparsity is essential for GAN training. The low-sparsity discriminator (blue lines) is too strong for the high-sparsity generator whereas the high-sparsity discriminator (green lines) does not have enough capacity to match the results of the dense model. Specifically, GAN training with an extremely sparse generator (i.e., $s_G=95\%, 90\%$) is a daunting task whose FID raises fastest due to the poor expressibility of the sparse generator. In the rest of the paper, we mainly focus on improving the trainability of this daunting but tantalizing setting.

\begin{algorithm*}[t]
\small
\caption{\small Sparse Training Unbalanced GAN (STU-GAN)}
\label{alg:STU-GAN}
{\bf Inputs:} 
Discriminator $D(x, \bm \theta_D)$, Generator $G(z, \bm \theta_G)$ , Sparsity of generator $s_G$, Sparsity of discriminator $s_D$, initial pruning rate $p$, exploration frequency $\Delta T$. \\
{\bf Output:} 
Sparse Generator $G({\bm z}, {\bm \theta_{s_{G}}})$, Sparse discriminator $D({\bm x}, {\bm \theta_{s_{D}}})$
\begin{algorithmic}[1] 

\State{$D({x}, {\bm \theta_{s_{D}}})$ $\gets$ $\mathrm{ERK}(D({ x}, \bm \theta), s_D$)} \Comment{Sparse initialization of $\mathrm{D}(x, \bm \theta_D)$}
\State{$G({z}, {\bm \theta_{s_{G}}})$ $\gets$ $\mathrm{ERK}(G({z}, \bm \theta), s_G$)} \Comment{Sparse initialization of $\mathrm{G(z, \bm \theta_G)}$}

\For {$t=1,2,\cdots$}
\State{${\bm \theta_{{s_{D}}, t}}={\rm Adam}\big(\nabla_{\bm \theta_{s_{D}}}\,\mathcal{L}_{GAN}({\bm \theta_{{s_{D}}, t-1}}, {\bm  \theta_{{s_{G}}, t-1})}\big)$}    
\State{${\bm \theta_{{s_{G}}, t}}={\rm Adam}\big(\nabla_{\bm \theta_{s_{G}}}\,\mathcal{L}_{GAN}({\bm \theta_{{s_{D}}, t}}, {\bm  \theta_{{s_{G}}, t-1})}\big)$}     

\If{($t$ \text{mod} $\Delta T$) == 0}
\State ${\bm{\mathrm{\theta}}_{s_{G}, t}^\prime} = \text{TopK}(\bm{\mathrm{\theta}}_{s, t},~1-p)$          \Comment{Parameter exploration of $G({\bm z}, {\bm \theta_{s_{G}}})$ using Eq.~\ref{eq:prune} and~\ref{eq:regrow}}    

\State ${\bm{\mathrm{\theta}}_{s_{G}, t} = \bm{\mathrm{\theta}}_{s_{G}, t}^\prime + \Phi(\bm{\mathrm{\theta}}_{i \notin \bm{\mathrm{\theta}}_{s_{G}, t}^\prime},~p)}$
\EndIf
\EndFor 
\end{algorithmic}
\end{algorithm*}

\section{Sparse Training Unbalanced GAN (STU-GAN)}
We have already known the challenge of training sparse GANs from scratch with the unbalanced sparsity distribution between the generator and discriminator. In this section, we introduce  Sparse Training Unbalanced GAN (\textbf{STU-GAN}) to address this challenge. The pseudocode of STU-GAN is detailed in Algorithm~\ref{alg:STU-GAN}.
Instead of training the generator with  static sparse connectivities, STU-GAN dynamically explores the parameter space spanned over the generator throughout training, enhancing the capacity of the highly sparse generator progressively. With the upgraded generator, STU-GAN mitigates the training instability of sparse unbalanced GANs. The parameter exploration is achieved by a prune-and-redistribute scheme, which enables increasing the effective weight space of the generator without increasing its parameter count.

Compared with the existing GAN compression techniques, the novelty of STU-GAN is located in: (1) STU-GAN directly trains a sparse GAN from scratch, and thus doesn't require any expensive pre-training steps; (2) STU-GAN starts from a sparse model and maintains the sparsity throughout training, making the approach more suited for edge devices. Specifically, the training process of STU-GAN comprises three main components: sparse initialization, sparse connectivity exploration, and sparse weight optimization, as explained below.

\subsection{Sparse Initialization} The choice of the layer-wise sparsity ratios (sparsity of each layer) plays an crucial role for sparse training~\cite{liu2022the}. Given that the most the state-of-the-art GANs are constructed based on convolutional neural networks (CNNs), we initialize both $G(z)$ and $D(x)$ with the \textit{Erd{\H{o}}s-R{\'e}nyi-Kernel} (ERK) graph topology~\cite{evci2020rigging}, which automatically allocates higher sparsity to larger layers and lower sparsity to smaller ones. Precisely, the sparsity of each CNN layer $l$ is scaled with $1 - \frac{n^{l-1}+n^{l}+w^{l}+h^{l}}{n^{l-1}\times n^{l}\times w^{l}\times h^{l}}$, where $n^l$ refers to the number of neurons/channels of layer $l$; $w^l$ and $h^l$ are the width and the height of the convolutional kernel in layer $l$. For non-CNN layers, the sparsity is allocated with  \textit{Erd{\H{o}}s-R{\'e}nyi} (ER)~\cite{mocanu2018scalable}  with $1 - \frac{n^{l-1}+n^{l}}{n^{l-1}\times n^{l}}$. ER and ERK  typically achieve better performance on CNNs than the naive uniform distribution, i.e., allocating the same sparsity to all layers~\cite{gale2019state}.

\subsection{Sparse Parameter Exploration} 
STU-GAN differs from the Static Sparse GAN training mainly in sparse parameter exploration. Sparse parameter exploration performs prune-and-redistribute to search for better sparse connectivity in generators, which in turn promotes benign competition of the minimax game by improving the quality of counterfeits. Concretely, after every $\Delta T$ iteration of training, we eliminate  percentage $p$ of parameters with the smallest magnitude from the current sparse subnetwork: 
\begin{equation}
    \bm{\mathrm{\theta}}_{s, t}^\prime = \text{TopK}(\bm{\mathrm{\theta}}_{s, t},~(1-p)\cdot N),
    \label{eq:prune}
\end{equation}
where $\mathrm{TopK}(v, k)$ returns the weight tensor retaining the $\rm top$-$\rm k$ elements from $v$; $\bm{\mathrm{\theta}}_{s, t}$ refers to the sparse subnetwork at $t$ training step; $N$ is the total number of parameters in sparse networks. $\bm{\mathrm{\theta}}_{s, t}^\prime$ is the set of parameters remaining after pruning. While magnitude pruning is simple, we empirically find that it performs better than other more complex pruning criteria such as connectivity sensitivity~\cite{lee2018snip}, gradient flow~\cite{Wang2020Picking}, and taylor expansion~\cite{molchanov2016pruning}, in the context of sparse training. 

To explore new parameters while maintaining the parameter count fixed, we redistribute the same number of pruned parameters back after pruning by:
\begin{equation}
    \bm{\mathrm{\theta}}_{s, t} = \bm{\mathrm{\theta}}_{s, t}^\prime + \Phi(\bm{\mathrm{\theta}}_{i \notin \bm{\mathrm{\theta}}_{s, t}^\prime},~p\cdot N)
    \label{eq:regrow}
\end{equation}
where function $\Phi(v, k)$ refers to growing $k$ weights picked from $v$ based on some certain criterion. $\bm{\mathrm{\theta}}_{i \notin \bm{\mathrm{\theta}}_{s, t}^\prime}$ are the zero elements located in $\bm{\mathrm{\theta}}_{s, t}^\prime$. More concretely, we choose the zero parameters with the largest gradient magnitude, which indicates the fastest loss reduction in the next iteration. The newly activated weights are initialized as zero to eliminate the historical bias.

This prune-and-redistribute scheme performs every $\Delta T$ training steps of the generator until s well-performing generator is converged at the end of the training process, corresponding to the situation where nearly all the parameters of the dense network have been activated. Similar to the synaptic pruning phenomenon~\cite{Chechik98synapticpruning,Chechik98neuronalregulation,Craik06cognitionthrough} in biological brains where some connections are strengthened while others are eliminated to learn new experiences, the prune-and-redistribute scheme allows the sparse connectivity pattern evolving during training to enhance the sparse generators.

\subsection{Sparse Weight Optimization} 
\textbf{Sparse Exponential Moving Averages (SEMA).} While exponential moving averages (EMA) has been widely used in various GANs models with strong results~\cite{karras2017progressive,yaz2018the,gidel2018a,Mescheder2018ICML}, it becomes less suited for STU-GAN. Since the newly activated weight has no historical information, the original update of EMA $\theta^{\rm EMA}_t = \beta \theta^{\rm EMA}_{t-1} + (1-\beta)\theta_{t}$ ends up with $\theta^{\rm EMA}_t = (1-\beta)\theta_{t}$. The large decay factor (e.g., $\beta=0.999$) immediately brings the newly activated weights close to zero. To address this problem, we proposed the sparse variant of EMA, SEMA, as following:
\begin{equation}
\label{Eq:SEMA}
\theta^{\rm SEMA}_{s,t}=\left\{
\begin{array}{lcl}
0  &   & {if \; T=0}, \\
\theta_{t}  &   & {if \; T=1}, \\
\beta \theta^{\rm SEMA}_{s, t-1} + (1-\beta)\theta_{t}  &   & {if \; T>1}. \\
\end{array}
\right.
\end{equation}
where $T$ refers to the total number of iterations where the weight $\theta$ has most recently been activated. In short, SEMA initializes the new activated weights as its original value $\theta^{t}$ instead of $ (1-\beta)\theta^{t}$. Except for this, the activated parameters are optimized with Adam~\cite{kingma2014adam} in the same way as the dense GANs. The zero-valued parameters are forced to be zero before the forward pass and after the backward pass to eliminate their contributions to the loss function.

\section{Experimental Results}

\label{sec:exp}
\textbf{Experimental Setup.} In this section, we conduct experiments to evaluate STU-GAN. Following ~\cite{chen2021gans}, we choose the widely-used SNGAN~\cite{miyato2018spectral} on CIFAR-10 for the image generation task. Moreover, to draw a more solid conclusion with large scale GANs, we also evaluate our method with BigGAN~\cite{brock2018large} trained on CIFAR-10 and ImageNet. To enable comparison among different methods, we follow~\cite{chen2021gans} and employ two widely-used metrics Fr\'echet Inception Distance (FID) and Inception Score (IS) as the approximate measure of model performance.

We set the exploration frequency as $\Delta T=500$ for BigGAN and $\Delta T=1000$ for SNGAN based on a small random search. The initial pruning rate of weight exploration is $p=0.5$ for all models, following~\cite{evci2020rigging,liu2021we}. The original hyperparameters (training epochs, batch size, etc.) and training configurations of GANs are the same as the ones used to train dense GANs\footnote{The training configurations and hyperparameters of BigGAN and SNGAN are obtained from the open-source implementations \url{https://github.com/ajbrock/BigGAN-PyTorch} and \url{https://github.com/VITA-Group/GAN-LTH}, respectively.}. 

\begin{figure*}[ht]
\begin{center}
\centerline{\includegraphics[width=0.95\textwidth]{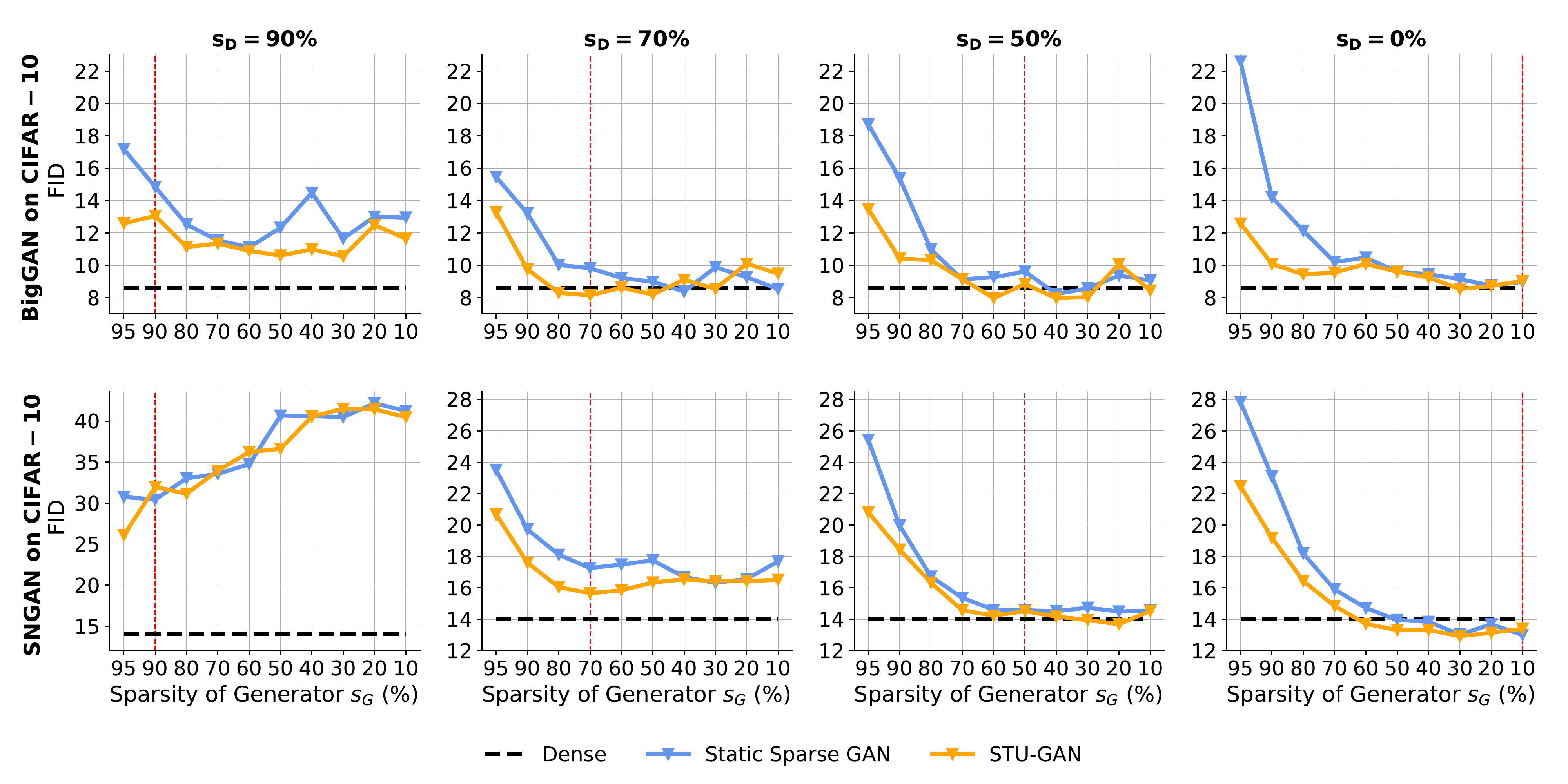}}
\caption{Comparisons between Static Sparse GAN and STU-GAN with various combinations between $s_G$ and $s_D$. The sparsity balanced setting with $s_D=s_G$ is indicated with dashed red lines.}
\label{Fig:static_vs_stu}
\end{center}
\end{figure*}

\subsection{Performance Improvement of STU-GAN over Static Sparse GAN} 
\label{sec:over_static}
The most direct baseline of STU-GAN is Static Sparse GAN, i.e., training a sparse GAN from scratch without any parameter exploration. We implement this baseline in the way that the only difference between STU-GAN and Static Sparse GAN is the sparse parameter exploration. The comparison is conducted at four settings with various sparsity levels in the discriminator $s_D\in[0.0, 0.5, 0.7, 0.9]$. We do so in the hope of providing insights into GAN training from the perspective of different sizes of $G(z)$ and $D(x)$ in terms of sparsity. The results are shown in Figure~\ref{Fig:static_vs_stu}. We see that STU-GAN consistently outperforms Static Sparse GAN with different settings, especially when trained with extremely sparse generators (i.e., higher $s_G$). Besides this, we further have the following observations:

\textbf{1. Possibility of improving GAN performance, rather than compromising it.} Applying STU-GAN on BigGAN with a sufficient sparse $D(x, \bm \theta_{s_D})$ (i.e., $s_D=50\%, 70\%$) can easily yield sparse generators with 20\% $\sim$ 90\% weights remained while achieving equal or even lower FID compared with the dense GANs (we call such subnetworks matching subnetworks). Impressively, the combination of a 80\% sparse generator and a 70\% sparse discriminator is good enough to surpass the performance of the dense GANs. This observation uncovers a very promising finding, that is, instead of training a dense GAN and then compressing it, directly training sparse GANs from scratch can be a better solution when the source budget is not extremely strict, i.e., allowing around 20\% $\sim$ 50\% weights remaining.

\textbf{2. STU-GAN substantially stabilizes sparse GAN training, even in the most unbalanced case.} Naively training with an extremely sparse generator and a dense discriminator is nearly impossible due to the notorious training instability of GANs. The generator is too weak to generate high-quality fake data. For instance, under scenarios with extremely sparse generators (i.e., $s_G=95\%$, first point of every line), performance achieved by Static Sparse GAN degrades significantly as discriminators get stronger from $s_D=70\%$ to $s_D=50\%$, and to $s_D=0\%$. In contrast, STU-GAN substantially improves the trainability of extremely unbalanced GANs with all $s_D$. Without using any pre-trained knowledge, STU-GAN decreases FID of BigGAN in the worst case ($s_D=0\%$ and $s_G=95\%$) from 22.5 to 12.5, while producing an extremely efficient generator with only 5\% parameters. 

\textbf{3. SNGAN is more sensitive to unbalance sparsity than BigGAN.} SNGAN suffers more from performance loss compared with BigGAN. As discriminators gets sparser, SNGAN has a sharp performance drop. With highly sparse discriminator ($s_D=90\%, 70\%$), SNGAN can not find any matching subnetworks. This is likely due to the unbalanced architecture design of SNGAN, where the model width (number of filters) of generators is twice larger than discriminators. Over-sparsifying discriminators would further amplify such parameter unbalance, leading to inferior performance.

\begin{table*}[h]
\scriptsize
\centering
\caption{(FID ($\downarrow$), IS ($\uparrow$)) of sparse BigGAN and SNGAN on CIFAR-10. We divide sparse methods into two groups, post-training pruning (PF and GAN Tickets) and sparse training (Static Sparse GAN and STU-GAN). ``$s_D (\%)$'' and ``$s_G (\%)$'' refers to the sparsity of the discriminator and the generator used to train GANs, respectively. Results of GAN Tickets are obtained from~\cite{chen2021gans}. The exact sparsity of GAN Tickets at $s_G=90\%$ is 87\%. Results are averaged from three training runs.}
\label{tab:com_cifar10}
\resizebox{\textwidth}{!}{
\begin{tabular}{l c ccc ccc}
\cmidrule[\heavyrulewidth](lr){1-8}

\textbf{Methods}  &   & \multicolumn{3}{c}{SNGAN} & \multicolumn{3}{c}{BigGAN}  \\ 

\cmidrule(lr){1-1}
\cmidrule(lr){2-2}
\cmidrule(lr){3-5}
\cmidrule(lr){6-8}
 Sparsity   & $s_D (\%)$ & $s_G=95\%$ & $s_G=90\%$ &$s_G=80\%$  & $s_G=95\%$  & $s_G=90\%$ &$s_G=80\%$      \\ 

\cmidrule(lr){1-1}
\cmidrule(lr){2-2}
\cmidrule(lr){3-5}
\cmidrule(lr){6-8}

&  &  \multicolumn{3}{c}{(FID ($\downarrow$), IS ($\uparrow$))} &  \multicolumn{3}{c}{(FID ($\downarrow$), IS ($\uparrow$))}    \\

\cmidrule(lr){1-1}
\cmidrule(lr){2-2}
\cmidrule(lr){3-5}
\cmidrule(lr){6-8}

Dense GAN & 0 &  (14.01, 8.26) & (14.01, 8.26) & (14.01, 8.26)    &  (8.62, 8.90) & (8.62, 8.90) & (8.62, 8.90)   \\
\cmidrule(lr){1-1}
\cmidrule(lr){2-2}
\cmidrule(lr){3-5}
\cmidrule(lr){6-8}

PF (global) & 0 & (26.34, 7.31)  & (17.97, 7.85) &  (16.99, 8.03) & (18.48, 8.04)  & (13.01, 8.51) & (10.52, 8.66)\\

PF (uniform) & 0 & (27.35, 7.27)  &  (18.23, 7.76) &  (16.93, 8.04)  &  (16.32, 8.23) & (12.27, 8.50) & (9.43, 8.73)   \\

GAN Tickets  & 0  & n/a &  (19.29, 8.07)  &  (16.79, 8.16)    & n/a &  (9.87, 8.75) & (9.06, 8.87)   \\ 
\cmidrule(lr){1-1}
\cmidrule(lr){2-2}
\cmidrule(lr){3-5}
\cmidrule(lr){6-8}
Static Sparse GAN & 50 & (25.43, 7.58) & (19.97, 7.78)  & (16.71, 8.01)   & (18.70,  7.82)  &  (15.37, 8.12)  & (10.98, 8.47)\\   
Static Sparse GAN & 70 & (23.52, 7.56) & (19.73, 7.69)  & (18.11, 7.93)  & (15.46,  8.13)  &  (13.20, 8.32)  & (10.03, 8.64)\\   
STU-GAN (ours) & 50 & (22.34, 7.71) & (18.78, 7.94) & (16.32, 8.01)   & (13.47, 8.19) & (10.42, 8.57) & (10.34, 8.61)      \\
STU-GAN (ours) & 70 & \textbf{(20.67, 7.77)} & \textbf{(17.60, 8.07)} & \textbf{(16.04, 8.18)}  & \textbf{(13.27, 8.37)} & \textbf{(9.75, 8.80)} & \textbf{(8.57, 8.91)}      \\
\cmidrule[\heavyrulewidth](lr){1-8}
\end{tabular}}
\end{table*}

\begin{table*}[h]
\scriptsize
\centering
\caption{Training and testing FLOPs of sparse BigGAN and SNGAN on CIFAR-10. The FLOPs of sparse GANs are normalized with the FLOPs required to train and test dense GANs. The methods that require the smallest number of training FLOPs are marked as bold.}
\label{tab:com_cifar10_flops}
\resizebox{\textwidth}{!}{
\begin{tabular}{l c ccc ccc}
\cmidrule[\heavyrulewidth](lr){1-8}

\textbf{Methods}  &   & \multicolumn{3}{c}{SNGAN} & \multicolumn{3}{c}{BigGAN}  \\ 

\cmidrule(lr){1-1}
\cmidrule(lr){2-2}
\cmidrule(lr){3-5}
\cmidrule(lr){6-8}
 Sparsity   & $s_D (\%)$ & $s_G=95\%$ & $s_G=90\%$ &$s_G=80\%$  & $s_G=95\%$  & $s_G=90\%$ &$s_G=80\%$      \\ 

 \cmidrule(lr){1-1}
\cmidrule(lr){2-2}
\cmidrule(lr){3-5}
\cmidrule(lr){6-8}

&  &  \multicolumn{3}{c}{(Training FLOPs $\times10^{17}$, Testing FLOPs $\times10^{9}$)} &  \multicolumn{3}{c}{(Training FLOPs $\times10^{18}$, Testing FLOPs $\times10^{9}$)}    \\

\cmidrule(lr){1-1}
\cmidrule(lr){2-2}
\cmidrule(lr){3-5}
\cmidrule(lr){6-8}

Dense GAN & 0 &  (0.5, 3.36) & (0.5, 3.36) & (0.5, 3.36) & (2.05, 3.36) & (2.05, 3.36) & (2.05, 3.36)   \\
\cmidrule(lr){1-1}
\cmidrule(lr){2-2}
\cmidrule(lr){3-5}
\cmidrule(lr){6-8}

PF (global) & 0 & (1.06$\times$, 0.05$\times$)  & (1.11$\times$, 0.10$\times$) &  (1.21$\times$, 0.20$\times$) & (1.05$\times$, 0.05$\times$)  & (1.10$\times$, 0.10$\times$) &  (1.20$\times$, 0.20$\times$)\\

PF (uniform) & 0 & (1.05$\times$, 0.05$\times$)  & (1.10$\times$, 0.10$\times$) &  (1.20$\times$, 0.20$\times$) &  (1.05$\times$, 0.05$\times$)  & (1.10$\times$, 0.10$\times$) &  (1.20$\times$, 0.20$\times$) \\

GAN Tickets  & 0  & n/a &  (4.57$\times$, 0.13$\times$)  &  (4.16$\times$, 0.21$\times$)    & n/a &  (4.57$\times$, 0.13$\times$)  &  (4.17$\times$, 0.21$\times$)  \\ 
\cmidrule(lr){1-1}
\cmidrule(lr){2-2}
\cmidrule(lr){3-5}
\cmidrule(lr){6-8}

Static Sparse GAN & 50 & (0.23$\times$, 0.05$\times$) & (0.26$\times$, 0.10$\times$)  & (0.32$\times$, 0.20$\times$)   &  (0.37$\times$, 0.05$\times$) & (0.39$\times$, 0.10$\times$)  & (0.41$\times$, 0.20$\times$)  \\   
Static Sparse GAN & 70 & (\textbf{0.16$\times$, 0.05$\times$}) & (\textbf{0.19$\times$, 0.10$\times$})  & (\textbf{0.25$\times$, 0.20$\times$})  &  (\textbf{0.23$\times$, 0.05$\times$}) & (\textbf{0.25$\times$, 0.10$\times$})  & (\textbf{0.27$\times$, 0.20$\times$}) \\   
STU-GAN (ours) & 50 & (0.23$\times$, 0.05$\times$) & (0.26$\times$, 0.10$\times$)  & (0.32$\times$, 0.20$\times$)  & (0.37$\times$, 0.05$\times$) & (0.39$\times$, 0.10$\times$)  & (0.41$\times$, 0.20$\times$)       \\
STU-GAN (ours) & 70 & (\textbf{0.16$\times$, 0.05$\times$}) & (\textbf{0.19$\times$, 0.10$\times$})  & (\textbf{0.25$\times$, 0.20$\times$})  &  (\textbf{0.23$\times$, 0.05$\times$}) & (\textbf{0.25$\times$, 0.10$\times$})  & (\textbf{0.27$\times$, 0.20$\times$}) \\   
\cmidrule[\heavyrulewidth](lr){1-8}
\end{tabular}}
\end{table*}

\subsection{Comparison with Stronger Baselines}
\label{sec:com_strong_baselines}
We compare STU-GAN with various strong baselines including Static Sparse GAN, pruning and fine-tuning (PF~\cite{han2015deep}), and GAN Tickets~\cite{chen2021gans}. Static Sparse GAN refers to naively training sparse unbalanced GAN with a fixed sparse pattern, whereas PF and GAN Tickets are two post-training pruning methods that operate on a pre-trained model. GAN Tickets is the recently proposed strong baseline, which discovers matching subnetworks by adopting iterative pruning on a fully trained dense GAN. Hence, the discriminator's sparsity of PF and GAN Tickets for training is $s_D=0\%$. We set the number of training epochs of STU-GAN and Static Sparse GAN the same as the dense GANs. As shown in Table~\ref{tab:com_cifar10}, STU-GAN consistently achieves the best performance with only 30\% parameters remaining in the discriminators. Even though trained from scratch, STU-GAN outperforms those methods that require expensive iterative pruning and re-training, highlighting the superiority of our method in the trade-off between performance and efficiency.

\subsection{Computational Savings}
To evaluate the claimed computational savings achieved by our method, we report the training and testing FLOPs of all the methods that we consider in Section~\ref{sec:com_strong_baselines}. Training FLOPs refer to the overall FLOPs to obtain the final sparse generator, including computations of pre-training, re-training, and sparse training. Testing FLOPs represent the computations of the obtained generator to generate one single image. Results are summarized in Table~\ref{tab:com_cifar10_flops}. As expected, all the post-training pruning methods require more training FLOPs than training a dense model. In particular, GAN Tickets is the most power-hungry approach, whose total training FLOPs is 4.57$\times$ larger than dense training. Contrariwise, sparse-to-sparse training demonstrates great potential to accelerate training, saving 84\%, 81\%, and 75\% training FLOPs of SNGAN when training with a $95\%$, $90\%$, and $80\%$ sparse generator from scratch, respectively. STU-GAN further boosts the performance over Static Sparse GAN without any additional overhead for either training or testing.

\subsection{Analysis}
\begin{figure*}[h]
\begin{center}
\centerline{\includegraphics[width=0.9\textwidth]{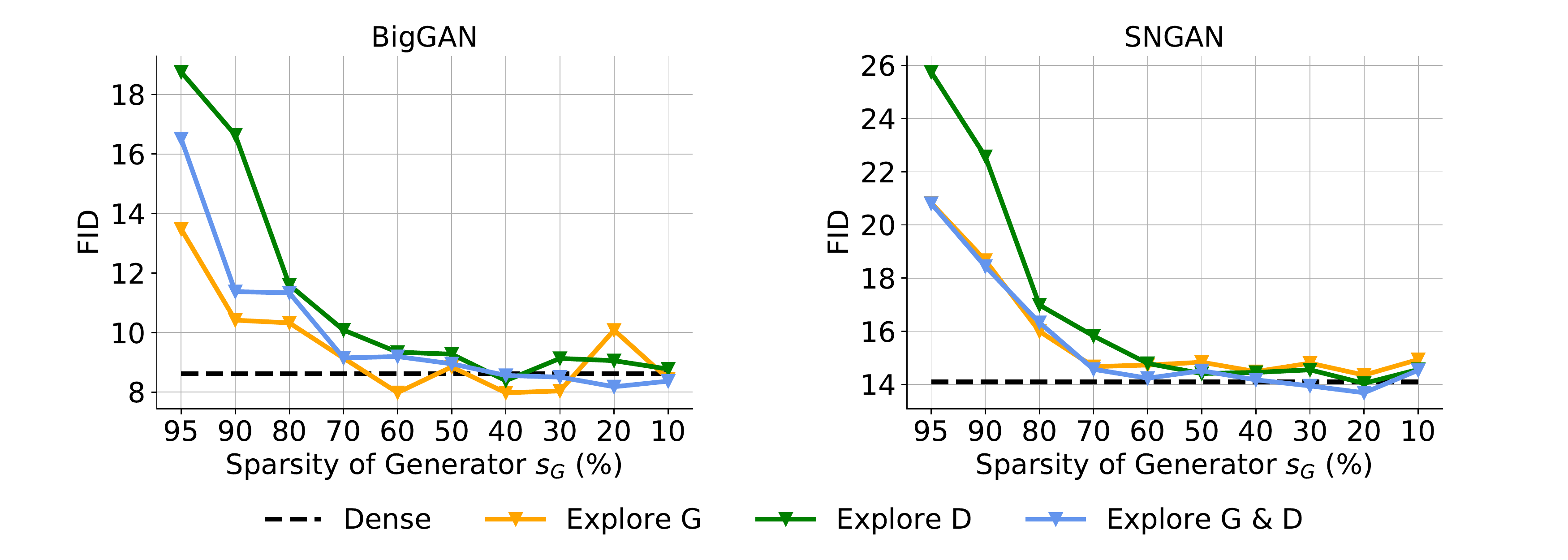}}
\caption{Effect of parameter exploration on different components. ``Explore G'', ``Explore D'', and ``Explore G \& D'' refers to applying parameter exploration only to generators, discriminators, and both, respectively. Experiments are conducted with 50\% sparse discriminators, i.e., $s_D=50\%$.}
\label{Fig:explore_GD}
\end{center}
\end{figure*}

\begin{figure*}[h]
\begin{center}
\centerline{\includegraphics[width=0.90\textwidth]{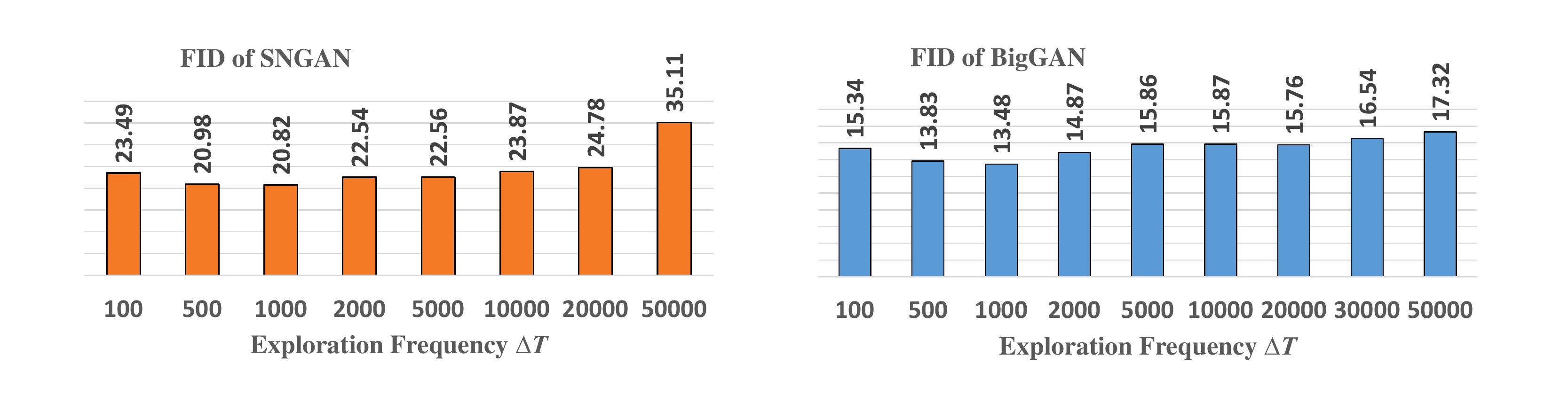}}
\caption{FID of sparse unbalanced BigGAN and SNGAN on CIFAR-10 with various $\Delta T$. The results of BigGAN in trained with 95\% sparse $D(x, \bm \theta_{s_{D}})$  and 95\% sparse $G(z, \bm \theta_{s_{G}})$.  The results of SNGAN in trained with 50\% sparse $D(x, \bm \theta_{s_{D}})$ and 95\% sparse $G(z, \bm \theta_{s_{G}})$. }
\label{Fig:exploration_fre}
\end{center}
\end{figure*}

\begin{table*}[!ht]
\footnotesize
\centering
\caption{(FID ($\downarrow$), IS ($\uparrow$)) of sparse BigGAN and SNGAN on CIFAR-10 with 5$\times$ training steps.}
\label{tab:ITOP}
\begin{tabular}{l c ccc ccc}
\cmidrule[\heavyrulewidth](lr){1-8}
\textbf{Methods}  &  & \multicolumn{3}{c}{SNGAN} &  \multicolumn{3}{c}{BigGAN}\\
\cmidrule(lr){1-1}
\cmidrule(lr){2-2}
\cmidrule(lr){3-5}
\cmidrule(lr){6-8}
  Sparsity  & $s_D (\%)$  & $s_G=95\%$ & $s_G=90\%$ &$s_G=80\%$ & $s_G=95\%$ & $s_G=90\%$ &$s_G=80\%$ \\
\cmidrule(lr){1-1}
\cmidrule(lr){2-2}
\cmidrule(lr){3-5}
\cmidrule(lr){6-8}
STU-GAN & 50  & (22.34, 7.71) & (18.78, 7.94) & (16.32, 8.01) & (13.47, 8.19) & (10.42, 8.57) & (10.34, 8.61)   \\
STU-GAN$_{5\times}$ & 50 &  \textbf{(18.44, 7.97)} & \textbf{(16.96, 7.98)} & \textbf{(15.68, 8.04)} & \textbf{(12.92, 8.44)} & \textbf{(10.01, 8.62)} & \textbf{(9.84, 8.71)} \\
\cmidrule[\heavyrulewidth](lr){1-8}
\end{tabular}
\end{table*}
\subsubsection{Which Components Should We Explore}
We have learned that parameter exploration in sparse generators improves the expressibility of generators and stabilizes the training of sparse unbalanced GANs with extreme sparse generators. Since parameter exploration is a universal technique that can potentially improve the expressibility of all sparse networks, we expect that its application on relatively stronger discriminators (i.e., $s_D=50\%$) will further amplify the training instability of sparse unbalance GANs, leading to counter-productive results. We evaluate our expectation in Figure~\ref{Fig:explore_GD}. The results are on par with our expectations. Simultaneously exploring parameters in $G(z, \bm \theta_{s_{G}})$ and $D(x, \bm \theta_{s_{D}})$ achieves higher (i.e., BigGAN) or equal (i.e., SNGAN) FID than solely exploring $G(z, \bm \theta_{s_{G}})$. However, solely exploring parameters of $D(x, \bm \theta_{s_{D}})$, as we expected, amplifies the training stability and increases FID significantly.

\subsubsection{Effect of Exploration Frequency}
The exploration frequency $\Delta T$ (i.e., the number of training steps between two iterations of parameter exploration) directly controls the trade-off between the quality and quantity of the parameter exploration in STU-GAN. Smaller $\Delta T$ means more parameter exploration, by extension, more exploration of the parameter space. On the contrary, larger $\Delta T$ allows the subnetworks between two explorations to be well-trained, improving the correctness of the parameter exploration. We report the trade-off in Figure~\ref{Fig:exploration_fre}. We see that the best performance is obtained when $\Delta T$ is set around 1000. BigGAN seems to be more robust to the exploration frequency compared with SNGAN.

\subsubsection{Boosting STU-GAN with In-Time Over-Parameterization} In-Time Over-Parameterization (ITOP) was introduced in~\cite{liu2021we} to understand the success of DST. ITOP points out that the off-the-shell DST methods, intentionally or unknowingly, all perform a process of ``over-parameterization” during training by gradually exploring the parameter space spanned over the model, and the performance of sparse training is highly correlated with the overall number of explored parameters during training. Here, we want to test if exploring more parameters of the generators can lead to better generative ability. More specifically, we extend training steps by 5 times while keeping the exploration frequency the same, such that we can perform more parameter exploration. We report the results in Table~\ref{tab:ITOP}. We see that the longer training time brings consistent performance gains to sparse GAN training, suggesting a possible way to further improve the performance of STU-GAN.

\subsubsection{Performance on ImageNet} To draw more solid conclusions, we evaluate STU-GAN with BigGAN on ImageNet, a complex dataset with high-resolution $128\times 128$ and diverse samples. We compare STU-GAN with fully dense BigGAN and a small dense BigGAN with half of the filters in the generator and the discriminator (termed Small Dense BigGAN). We train sparse BigGAN with STU-GAN at sparsity of $s_G \in [90\%, 60\%]$ and $s_D=50\%$ for comparison. The results are shown in Table~\ref{tab:imagenet}. Again, we find that STU-GAN is able to outperform the performance of fully dense BigGAN on ImageNet with a 50\% sparse discriminator and a 60\% sparse generator. Moreover, with even a smaller parameter count, our method improves FID of the small dense model by 3.88 FID. Very impressively, STU-GAN can still match the performance of the small dense model when only 10\% parameters are remained in the generator, highlighting a more appealing direction to allocate limited parameter budgets.

\begin{table}[!h]
\footnotesize
\centering
\caption{FID ($\downarrow$) of BigGAN on ImageNet 128$\times$128 without the truncation trick. }
\label{tab:imagenet}
\begin{tabular}{l|ccc}
\toprule
\textbf{Methods} & $s_D (\%)$ & $s_G (\%)$ & FID ($\downarrow$)  \\ 
\midrule
Dense BigGAN & 0 & 0 & 9.81 \\
Small Dense BigGAN  &  50 & 50 & 13.56  \\
\midrule
STU-GAN & 50 & 60 & \textbf{9.68} \\
STU-GAN & 50 & 90 & 13.39  \\
\bottomrule
\end{tabular}
\end{table}

\section{Conclusion}\label{sec13}

In this paper, we study GAN training from the perspective of sparsity. We demonstrate that the well-matched sparsity between generators and discriminators is essential to GAN training, whereas the sparsity unbalanced scenarios significantly degrade the trainability of sparse GNA. We further explore the possibility of training sparsity-unbalanced GAN with an extremely sparse generator and a much denser discriminator by proposing Sparse Training Unbalanced GAN (STU-GAN). Training and maintaining only a small fraction of parameters without involving any pre-training, STU-GAN can outperform several strong after-training pruning techniques as well as the fully dense GANs, shedding light on
the appealing prospect of sparsity to stabilize and accelerate GAN training.

\backmatter


\bmhead{Acknowledgments}

This work is supported by Science and Technology Innovation 2030 –“Brain Science and Brain-like Research” Major Project (No. 2021ZD0201402 and No. 2021ZD0201405).

\bibliography{sn-bibliography}

\clearpage
\begin{appendices}

\begin{figure*}[h]
\begin{center}
\centerline{\includegraphics[width=0.6\textwidth]{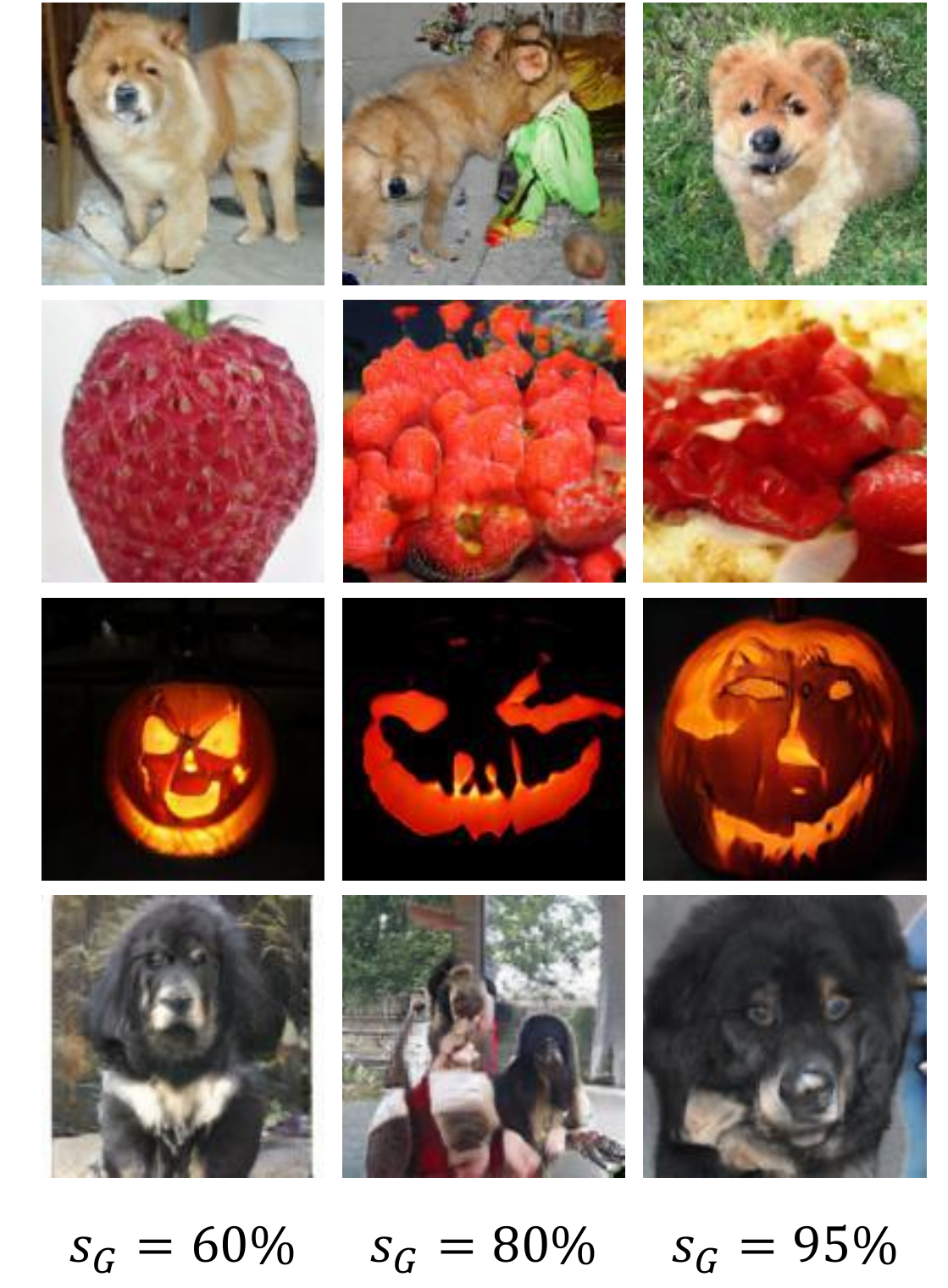}}
\caption{Example Images Generated by STU-GAN on ImageNet 128 $\times$ 128. Each row of images are generated via STU-GAN models with various $s_{G}$ and $s_D=50\%$.}
\label{Fig:generate_samples}
\end{center}
\end{figure*}




\end{appendices}

\end{document}